\documentclass[lettersize,journal]{IEEEtran}
\usepackage{amsmath,amsfonts}
\usepackage{algorithmic}
\usepackage{algorithm}
\usepackage{array}
\usepackage[caption=false,font=normalsize,labelfont=sf,textfont=sf]{subfig}
\usepackage{textcomp}
\usepackage{stfloats}
\usepackage{url}
\usepackage{verbatim}
\usepackage{graphicx}
\usepackage{cite}
\usepackage{diagbox}

\usepackage{newfloat}
\usepackage{listings}
\usepackage{multirow}
\usepackage{color}
\usepackage{amssymb}
\usepackage{booktabs}
\usepackage[colorlinks,urlcolor=blue,linkcolor=blue,citecolor=blue]{hyperref}
\newcommand{\cut}[1]{}

\begin{document}

\title{Self-Supervised Exploration via Temporal Inconsistency in Reinforcement Learning}

\author{Zijian Gao, Kele Xu, Yuanzhao Zhai, Bo Ding, Dawei Feng, Xinjun Mao, Huaimin Wang
		\thanks{This work was partially supported by the major Science and Technology Innovation 2030 ``New Generation Artificial Intelligence'' project 2020AAA0104803, and National Natural Science Foundation of China (No.62206307). (Corresponding author: Kele Xu.)}
		\thanks{Zijian Gao, Kele Xu, Yuanzhao Zhai, Bo Ding, Dawei Feng, Xinjun Mao and Huaimin Wang are with the School of Computer, National University of Defense Technology, Changsha 410000, China.}
		\thanks{This work has been submitted to the IEEE for possible publication. Copyright may be transferred without notice, after which this version may no longer be accessible.}
		
}

\markboth{Journal of \LaTeX\ Class Files,~Vol.~14, No.~8, August~2021}%
{Shell \MakeLowercase{\textit{et al.}}: A Sample Article Using IEEEtran.cls for IEEE Journals}


\maketitle

\begin{abstract}
Under sparse extrinsic reward settings, reinforcement learning has remained challenging, despite surging interests in this field. Previous attempts suggest that intrinsic reward can alleviate the issue caused by sparsity. 
In this article, we present a novel intrinsic reward that is inspired by human learning, as humans evaluate curiosity by comparing current observations with historical knowledge.
Our method involves training a self-supervised prediction model, saving snapshots of the model parameters, and using nuclear norm to evaluate the temporal inconsistency between the predictions of different snapshots as intrinsic rewards. We also propose a variational weighting mechanism to assign weight to different snapshots in an adaptive manner. 
Our experimental results on various benchmark environments demonstrate the efficacy of our method, which outperforms other intrinsic reward-based methods without additional training costs and with higher noise tolerance.

\end{abstract}

\begin{IEEEkeywords}
Deep reinforcement learning, Intrinsic rewards, Exploration, Nuclear norm
\end{IEEEkeywords}

\section{Introduction}
\label{Introduction}
Over the past few years, Deep Reinforcement Learning (DRL) has made great progress on several challenging sequential decision-making tasks, such as Atari games~\cite{atari}, board games~\cite{game2} and videos games~\cite {game3}. Promising results in DRL rely on reasonable and dense external rewards, which can be difficult to obtain as manually designing a well-defined reward function for each task may not feasible~\cite{RND}. The complex design of external incentives and the scarcity of external rewards severely constrain the performance of many DRL methods.

\begin{figure}[t]
\begin{center}
\centerline{\includegraphics[width=\columnwidth]{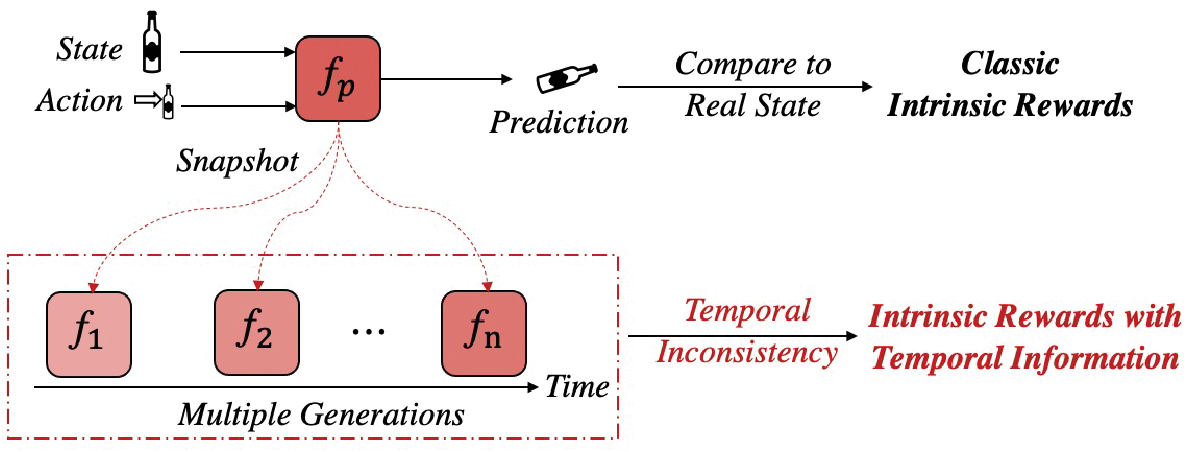}}
\caption{The schematic diagram of our method and classic prediction-based intrinsic reward. Classical intrinsic reward relies on a comparison between the prediction and the ground truth of next real state. Instead, we exploit temporal inconsistency across multiple generations for better self-supervised exploration.}
\label{diagram}
\end{center}
\end{figure}

For human agents, the intrinsic reward is a key factor in exploration, as it motivates people to seek and learn more without extrinsic reward ~\cite{burda2018large,ryan2000intrinsic,smith2005development}. Curiosity is a representative intrinsic reward for DRL and it has recently attracted increasing attention. Growing up, people are accustomed to limited external rewards with the self-driven motivation. Inspired by human curiosity, some works are proposed to design intrinsic rewards to compensate for the sparse extrinsic rewards in DRL~\cite{liu2021behavior,RND,Disagreemet}. Most intrinsic rewards can be divided into two broad categories: count-based methods and prediction-based methods. Count-based methods aim to maximize access to less explored states, which are not scalable to large-scale settings~\cite{bellemare2016unifying,10.5555/3172077.3172232,ostrovski2017count,lopes2012exploration,poupart2006analytic,RND                                                                                                                                                                                                                                                                                                                                                                                                                                                                                                                                                                            }. Prediction-based methods train a predictive model that estimates the next state based on state-action pairs, and evaluates the intrinsic reward by the difference between the predicted state and the ground truth state~\cite{ICM,Disagreemet,burda2018large,stadie2015incentivizing,active}. 

While potential applications for intrinsic reward seem obvious, the problem is still far from being solved. For example, many previous curiosity methods neglect temporal information. 
When a person encounters a problem, he will try to solve it by retrieving information from memory. If he fails, awareness of information discrepancy will lead to the arousal of curiosity~\cite{rotgans2017role}, indicating the importance of historical knowledge. As humans, curiosity is generated by not only using our intuitions about the current state but also employing our historical knowledge. 
Consider yourself a newborn baby in your parents' arms in a park. The erratic movement of the leaves is appealing at first, but your attention soon shifts to other, more exciting things. Because you accumulate historical information over a long period of observation and conclude that the movement of the leaves is meaningless and uninteresting.
Thus, without considering the historical generations of curiosity, it may be difficult to accurately evaluate the curiosity as it may be disrupted by noise. 

From an uncertainty perspective, stochasticity in the fitting process, including parameter initialization, exploration, replay sampling, and environment dynamics, leads to randomness in parameters and high-variance estimates~\cite{liang2022reducing}. As a result, existing curiosity-driven methods may be disrupted by noise, leading to inaccurate predictions and intrinsic rewards. Although prediction error and count error provide an estimate of uncertainty, they are insufficient and inaccurate due to the randomness and noise. To address this issue, utilizing historical knowledge to avoid high-variance estimates can provide more accurate intrinsic rewards.
\begin{figure}[t]
\begin{center}
\centerline{\includegraphics[width=\columnwidth]{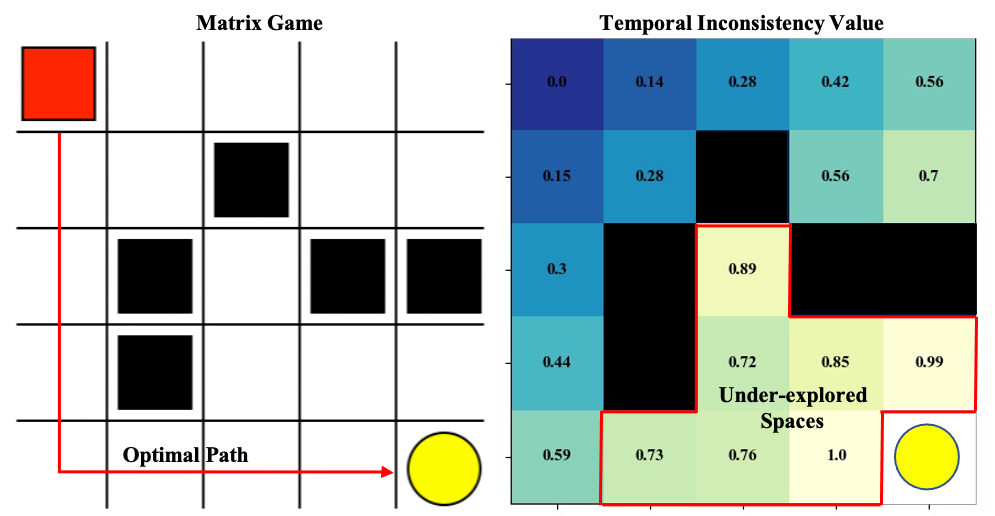}}
\caption{To illustrate our approach, we provide a toy example in the matrix game. Here, the red agent's objective is to reach the yellow circle while avoiding the black traps. We train a prediction model using the Q-learning algorithm and apply the sliding window strategy to save a set of snapshots of the prediction model during training. After training, we calculate the temporal inconsistency value between the snapshots and present the heat map of the normalized temporal inconsistency value in the right figure. The brightly colored areas indicate higher temporal inconsistency values. We observed that the temporal inconsistency value is much higher for under-explored spaces, which are more difficult for the agent to access, compared to the grids that the agent can access more easily.}
\label{toyexample}
\end{center}
\end{figure}

We introduce a Temporal Inconsistency-based Intrinsic Reward (TIR) that leverages the temporal information of the prediction model for better self-supervised exploration (Figure\ref{diagram}). We first train a prediction model based on state-action pairs to estimate the next state, and then use the sliding window strategy to save multiple snapshots of model parameters. We assume that under-explored state spaces should have higher temporal inconsistency values and intrinsic rewards. In the matrix game example (Figure\ref{toyexample}), we validate this hypothesis by normalizing the temporal inconsistency values and observing that grids where the agent frequently arrives have lower inconsistency values than grids where it rarely arrives. We use the nuclear norm to measure the inconsistency between the predictions of different snapshots, which serves as the intrinsic reward for DRL. Finally, we propose a variational self-weighted mechanism to improve performance by assigning weights to different snapshots.

In brief, our main contributions include: 

\begin{itemize}
    \item We propose a Temporal Inconsistency-based Intrinsic Reward (TIR), which employs the weighted nuclear norm to measure the distance between the predictions of the model's snapshots. The distances are then deployed as the intrinsic rewards for DRL.
   
    \item Furthermore, we design a variational weighting mechanism for different snapshots, with the goal to improve performance further while maintaining higher noise tolerance. The resulting reward is simple and straight-forward to implement, yet effective.
   
    \item We demonstrate that our proposed TIR outperforms competing methods on multiple benchmarks with different settings, including DeepMind Control Suite (DMC)\cite{dm_control} and Atari environments\cite{atari}. In Atari games, TIR shows overwhelming superiority over other competitive intrinsic-based methods. On DMC, we achieve state-of-the-art performance on 8 of 12 tasks across different domains, compared to both intrinsic reward-based baselines and other pre-training strategies.
\end{itemize}

The remainder of this article is organized as following. In Section~\ref{relatedwork}, we review the related work about the intrinsic reward and low-rank regularization. In Section~\ref{methodology}, we introduce a novel intrinsic reward method named TIR and the mechanism that enables the
agent to explore the environment by itself more efficiently. Further, in Section~\ref{experiment}, we show the performance of  and analyze the main components in Section~\ref{furtheranalysis}, which illustrates TIR can achieve the promising performance and robustness compared to strong baselines. The main conclusions are covered in Section~\ref{conclusion}.

\section{Related Work}
\label{relatedwork}
We will discuss the related work of intrinsic rewards and low-rank regularization in this section. We also talk about the differences and connections between our method and previous related works.
\subsection{Intrinsic Reward-based Exploration}
In DRL, many of previous exploratory attempts employ prediction error~\cite{ICM}, prediction uncertainty~\cite{Disagreemet}, or variants of uncertainty as self-driving motivations in the learning phase~\cite{dean2020see,houthooft2016curiosity-driven,still2012information}. Most intrinsic reward-based methods can be interpreted as exploring areas of high uncertainty, and curiosity is measured by estimating the deep learning model's uncertainty (confidence). Some attempts use count-based~\cite{kearns2002near,charikar2002similarity} or pseudo-count-based exploration~\cite{10.5555/3172077.3172232                                                                                                                                                                                                                                                                                                                                                                                                                                                                                                                                                                                                                                                                                                                                                                                                                                                                                                                                                                                                                                                                                                                                                                                                                                                                                                                                                                                      }, which struggle with the scalability to higher-dimensional state spaces~\cite{bellemare2016unifying,ostrovski2017count,10.5555/3172077.3172232}.
Motivated by counting, RND~\cite{RND} proposes to measure the gap between the predictor and the target as an intrinsic reward, which is easy to implement and can be more efficient.

Recently, curiosity via self-supervised prediction~\cite{ICM,Disagreemet,active,burda2018large} is one prominent way to self-supervised exploration. The prediction-based intrinsic reward is defined as the error between the predicted state and the ground truth of the next state. But accurate prediction can be difficult to obtain in practical settings. For this reason, Disagreement~\cite{Disagreemet} proposes to use the variance of predictions by ensembling multiple models' predictions, instead of using a single model. Deep ensemble~\cite{dietterich2000ensemble,wu2020deep} technique is widely used to estimate the uncertainty in DRL, which proposes to train an ensemble of deep models rather than a single model and achieves superior results but is also computationally expensive.
NNM~\cite{chen2022nuclear} utilizes nuclear norm instead of variance to evaluate intrinsic rewards more accurately. Inspired by NNM, we also evaluate the temporal inconsistency using the variational weighting mechanism based on nuclear norm to get a more accurate estimate. 

However, in both prediction model-based NNM and Disagreement, training multiple deep networks can be extremely time-consuming and may introduce more stochasticity due to different initial parameters, which may cause inaccurate intrinsic rewards. More importantly, prediction-based methods evaluate the intrinsic rewards by only using the current prediction model. Our method also employs a self-supervised exploration paradigm, but relies on the temporal inconsistency between multiple snapshots. The set of snapshots is essentially the deep ensemble on the temporal dimension and can avoid the variance introduced by stochastic multiple parameters without additional training costs. With multiple generations of the predictive model, the intrinsic reward can be more accurate and better encourage agents to investigate more novel states using the temporal information~\cite{liu2021behavior,urlb}.
\begin{figure}[t]
\begin{center}
\centerline{\includegraphics[width=\columnwidth]{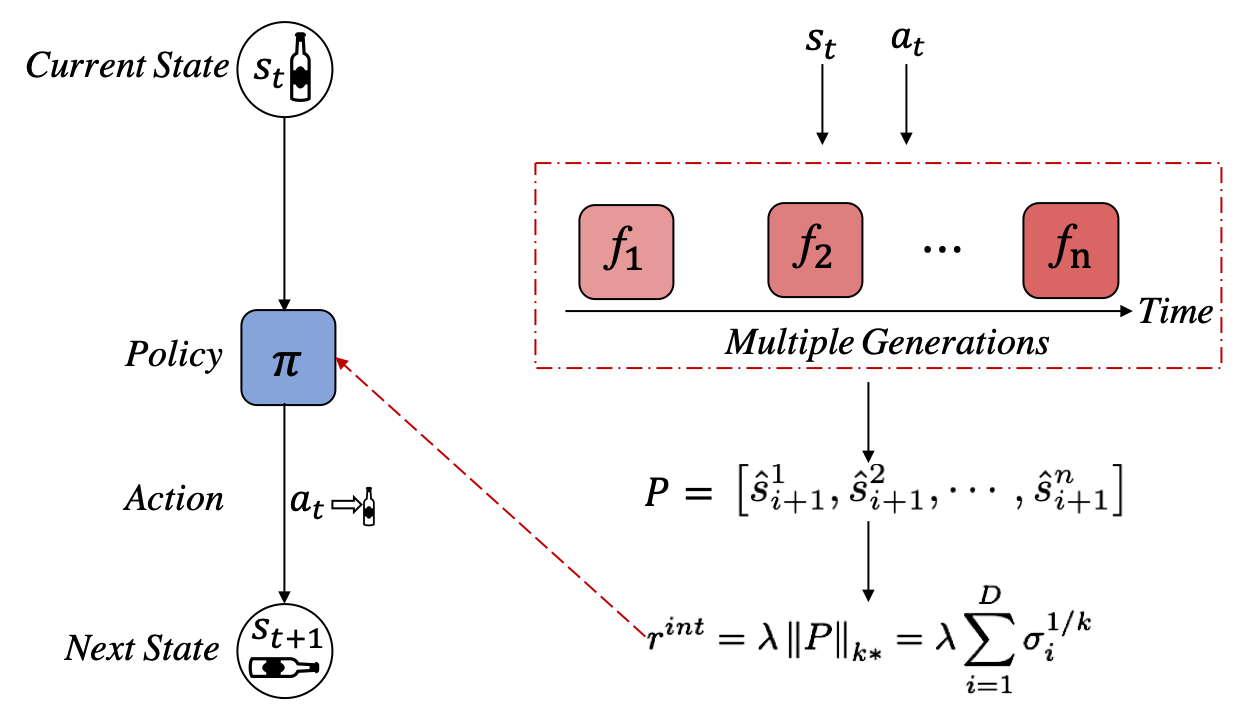}}
\caption{Overview structure of TIR. At time step $t$, the agent in the state $s_t$ takes action $a_t$ generated by policy $\pi$ and gets the next state $s_{t+1}$. We utilize multiple generations of the prediction model to get the $n$ prediction values as matrix $\mathbf{P}$, and calculate the variational weighting nuclear norm $\|\mathbf{P}\|_{k *}$ as the intrinsic reward $r_t^{int}$ to train policy $\pi$.}
\label{method}
\end{center}
\end{figure}
\subsection{Low-rank Regularization}
This subsection discusses the relationship between the proposed method and the low-rank regularization, as we utilize the weighted nuclear norm to measure the temporal inconsistency between multiple snapshots.
Low-rank matrices can be found in multiple fields, such as computer vision and recommendation system~\cite{li2020empirical,udell2019big}. Hu et al.~\cite{hu2021low} provides a comprehensive survey of low-rank regularization (LRR), focusing on the impact of using LRR as a loss function via nuclear norm.  Xiong et al.~\cite{xiong2021wrmatch} applies the nuclear norm to avoid prediction diversity collapse and improve the generalization ability of the classification model. The nuclear norm (NN) is a popular convex surrogate function of the $rank$ function widely used for low-rank regularization. 
Essentially, compared to the $\ell^2$ norm which is widely used in previous attempts to measure the distance, the nuclear norm not only encourages diversity, but also distinguishes irrelevance. In contrast to previous attempts, our work aims to maximize the weighted nuclear norm of the prediction matrix introduced by multiple snapshots of the prediction model.

\section{Methodology}
In this section, we first introduce the details of temporal inconsistency-based curiosity, then describe how to assign the weights to different snapshots via the proposed variational weighting mechanism.
\label{methodology}

We firstly train a prediction model to predict the next state $s_{t+1}$ with given state-action pairs $[s_t,a_t]$:
\begin{equation}
\hat{s}_{t+1}=f(s_t,a_t;\theta_p),
\end{equation}
where $\hat{s}_{t+1}$ is the prediction and the network parameter $\theta_p$ is optimized by minimizing the loss function $L_{p}$:

\begin{equation}
L_{p}=\frac{1}{2}\left\|\hat{s}_{t+1}-s_{t+1}\right\|_{2}^{2}.
\label{update}
\end{equation}

\subsection{Temporal Inconsistency-based Curiosity}

During the training phase, we save a snapshot of the model parameters along the optimization path, for every $j$ epoch. Therefore, as shown in Figure~\ref{method}, we can get $n$ models in temporal order, which can be denoted as $\{f_1,f_2,\cdots,f_n\}$. The sliding window strategy is used, and we replace the older model using the new one.
Given a state-action pair $[s_t,a_t]$, we can obtain a set of predicted values $\hat{s}_{t+1}$ for the next state using $n$ snapshots of the predictive model, then we use the prediction matrix $\mathbf{P}$ containing the predicted values to evaluate intrinsic rewards.

For a subset of the state space that the agent has been thoroughly explored, most of the snapshots can show similar prediction ability after sufficient sampling and training, so higher consistency can be obtained across multiple snapshots. For under-explored parts, the snapshots show high prediction errors and uncertainties because of insufficient training and different parameters, resulting in higher inconsistency over time along the temporal order. Obviously, the stronger the temporal inconsistency between predicted values is, the less exploration of state-action pairs can be, which has been validated in the toy example (Figure~\ref{toyexample}).
However, how to assess the temporal inconsistency of predicted values is also an under-explored topic.

\subsection{Measuring Inconsistency using Weighted Nuclear Norm}
To measure the distance between multiple vectors, different norms can be used, such as the $\ell_1$, $\ell_2$, $\ell_F$. Instead, based on our analysis proposed in \cite{chen2022nuclear}, we utilize the nuclear norm to measure the inconsistency between different snapshots, due to the superior performance and high noise tolerance. 
Moreover, we propose a novel weighting mechanism, with the goal to assign different weights to the snapshots. 

\begin{algorithm}[tb]
\small
 \caption{Temporal Inconsistency-based Intrinsic Rewards}
 \label{alg1}
 \textbf{Initialization}: policy network $\pi_\xi(a|s)$, prediction model $f_p$, the number of snapshots $N$, the training step counter $c$, the maximum episode step $T$, snapshot models $\{f_1,f_2,\cdots,f_n\}$, coefficient of intrinsic reward $\alpha$, coefficient of extrinsic reward $\beta$, coefficient $k_{ini}$, coefficient $k$, and interval $j$.
 \begin{algorithmic}[1] 
  \STATE $c = 0$
  \WHILE{$c <= 5e^7$}
  \FOR{$t=1,\cdots,T$}  
  \STATE $c = c + 1$
  \STATE  Receive observation $s_t$ from environment
  \STATE $a_t \leftarrow \pi_\xi(a|s)$ based on policy network $\pi_\xi$
  \STATE Take action $a_t$, receive observation $s_{t+1}$ and extrinsic reward ${r_{t}}^{ext}$ from environment
  \IF{the episode is finished}
  \STATE Store episodic data and end for
  \ENDIF
  \STATE $s_t\leftarrow s_{t+1}$
  \ENDFOR
  \STATE Sample batch data as $\left\{\left(s_{i}, a_{i}, {r_{i}}^{ext}, s_{i+1}\right)\right\}_{i=1}^{N}$from reply buffer
  \FOR{$each\,i = 1,\cdots,N$}
  \STATE Predict the next state $\hat{s}_{i+1}^{1} = f_{1}(s_i,a_i), \hat{s}_{i+1}^{2} = f_{2}(s_i,a_i)
        ,\cdots, \hat{s}_{i+1}^{n} = f_{n}(s_i,a_i)$
    \STATE Form matrix $\mathbf{P}$ by concatenating each state vector $\left[\hat{s}_{i+1}^{1}, \hat{s}_{i+1}^{2}, \cdots, \hat{s}_{i+1}^{n}\right]$
    \STATE Calculate intrinsic reward $r^{int}_i = \lambda\left\|\mathbf{P}\right\|_{k *}=\lambda\sum_{i=1}^{D} \sigma_{i}^{1 / k}.$
    \STATE Calculate total reward ${r_{i}}^{total}=\alpha {r_{i}}^{int} + \beta r^{ext}_i$
    \ENDFOR
    \STATE Update $f_{p}$ with sampled data by minimizing loss with Equation~\ref{update}
    \STATE Every j epochs, replace the earliest snapshot model with $f_p$
    
    \STATE $k = max(1, k_{ini} - \frac{u}{U}(k_{ini}-1))$  
    \STATE Update $\xi$ with sampled data by maximizing $r^{total}$ using RL algorithm
    
    \ENDWHILE
 \end{algorithmic}
\end{algorithm}

\subsubsection{Nuclear Norm based Intrinsic Reward}

For Disagreement~\cite{Disagreemet}, the inconsistency of prediction models can be measured by the variance of the predictions. Unlike the Disagreement method which trains multiple models in parallel, we only train one prediction model and get $n$ snapshots of the model for free, without incurring additional training costs. Concretely, the prediction matrix $\mathbf{P}$ consists of $n$ prediction from the snapshots and each prediction is a vector of $m$-dimension. Thus, $\mathbf{P}$ is with the size of $m\times n$. As each row of the matrix represents the probability distribution of the next state's prediction, higher inconsistency means greater differences between the rows of the matrix $\mathbf{P}$. Hence, a naive way to measure the prediction inconsistency is the matrix $rank(\mathbf{P})$, which can represent the linear irrelevance between the rows. Accordingly, we can generate our intrinsic reward by maximizing $rank(\mathbf{P})$, with the goal to encourage the agent to explore the states with higher temporal inconsistency.

However, directly maximizing the $rank(\mathbf{P})$ is an NP-hard, non-convex problem as the value of matrix rank is discrete. Thus, $rank(\mathbf{P})$ can not be employed directly for the DRL purpose. Moreover, due to the contamination of the noise, such as the environment stochasticity, the prediction values can be inaccurate which can incur additional noise. A flurry of studies attempts suggested that the matrix rank can be replaced by the nuclear norm, which is the convex envelope of $rank(\mathbf{P})$. Specifically, the nuclear norm can be formulated as:
\begin{equation}
\left\|\mathbf{P}\right\|_{*}=\operatorname{tr}\left(\sqrt{\mathbf{P}^{T} \mathbf{P}}\right)=\sum_{i=1}^{D} \sigma_{i},
\end{equation}
where $\operatorname{tr}(.)$ refers to the trace of the $\mathbf{P}$, and $D=\min(m, n)$. $\sigma_{i}$ denotes the $i_{th}$ largest singular value of $\mathbf{P}$. Mathematically, nuclear norm and Frobenius norm (F norm) are the boundaries of each other and the F-norm can be formulated as follows:
\begin{equation}
\left\|\mathbf{P}\right\|_{F}=\sqrt{\sum_{i=1}^{m} \sum_{j=1}^{n}\left|\mathbf{P}_{i, j}\right|^{2}}.
\end{equation}

According to~\cite{NN1,NN2}, the mutual boundary relationships
between $\left\|\mathbf{P}\right\|_{*}$ and $\left\|\mathbf{P}\right\|_{F}$ could be formulated as follows:
\begin{equation}
\frac{1}{\sqrt{D}}\|\mathbf{P}\|_{*} \leq\|\mathbf{P}\|_{F} \leq\|\mathbf{P}\|_{*} \leq \sqrt{D} \cdot\|\mathbf{P}\|_{F}.
\end{equation} 

Furthermore, Cui et al.~\cite{NN3} proved that $\left\|\mathbf{P}\right\|_{F}$ is strictly
opposite to Shannon entropy in monotony, and maximizing $\left\|\mathbf{P}\right\|_{F}$ is equal to minimizing entropy. So, the nuclear norm can assess not only the diversity but also distinguish irrelevance, which can be deployed to measure the inconsistency of the prediction matrix $\mathbf{P}$.

It is well-known that: the scale of reward value is also a key factor for DRL. Accordingly, to adapt the intrinsic reward to a suitable range, we introduce the parameter $\lambda$ to scale $\left\|\mathbf{P}\right\|_{*}$ to the same scale of the $\left\|\mathbf{P}\right\|_{F}$. Thus, we can define our intrinsic reward as:
\begin{equation}
    r^{int} = \lambda\left\|\mathbf{P}\right\|_{*},
\end{equation}
and the parameter $\lambda$ is set as 0.001 for all experiments.

\subsubsection{Variational Weighting Mechanism}

In our practical implementations, we find that the early snapshots of the model may be under-fitting, and produce inaccurate predictions. The inaccurate predictions will result in large singular values of the matrix, which is a perturbation for generating accurate intrinsic rewards.
However, the standard nuclear norm treats each singular value by the averaging weight, which leads to suboptimal performance. As a result, we propose an adaptive weight of the singular values to avoid the excessive influence of under-fitting snapshots. Here, a smaller weight will be assigned to the larger singular value. The weighted nuclear norm can be formulated as follows:
\begin{equation}
\|\mathbf{P}\|_{w *}=\sum_{i=1}^{D} w_{i} \sigma_{i},
\end{equation}
where $w_{i} \geq 0$ is a non-negative weight for $\sigma_{i}$. 

Nevertheless, in practice, it can be difficult to manually design the weight $w_{i}$ for each singular value in every specific scenario. Here, we propose an adaptive weighted nuclear norm according to singular values themselves:
\begin{equation}
\|\mathbf{P}\|_{k *}=\sum_{i=1}^{D} \sigma_{i}^{1 / k},
\end{equation}
where $k=\left\{2,3,…\right\}$ and $\sigma_{i}^{(1-k)/k}$ is treated as the weight for $\sigma_{i}$. It's obvious that in this formula the bigger $\sigma_{i}$ is, the smaller the weight is, which is in line with our principle. Besides that, the greater $k$, the more severe the inhibition on the large singular value. The adaptive weighted nuclear norm can avoid overestimation of intrinsic rewards due to insufficient prediction ability. However, as training continues, the predictive ability of the snapshots can be improved, which can provide more accurate predictions. As a result, the inhibition on large singular values should be reduced, and the parameter $k$ should be variable, gradually decreasing to 1 as the experiment progresses.

To conclude, the whole intrinsic reward can be formulated as the following equation, leveraging the variational weighting nuclear norm:
\begin{equation}
    r^{int} = \lambda\left\|\mathbf{P}\right\|_{k *}=\lambda\sum_{i=1}^{D} \sigma_{i}^{1/k}.
\end{equation}

Thus, we can get the optimization goal of the agent:
\begin{equation}
    \max_{\xi} \mathbb{E}_{\pi\left(S_{t} ; \xi\right)}\left[\sum \gamma^{t} (\alpha r_{t}^{int} + \beta r_{t}^{ext})\right],
\end{equation}
where $\gamma$ is the discount factor and $\xi$ represents parameters of policy $\pi$, $\alpha$, and $\beta$ are the coefficient of intrinsic reward and extrinsic reward respectively. Figure~\ref{method} and Algorithm~\ref{alg1} present the whole framework and pseudo-code of TIR. 

\section{Experimental Results}
\label{experiment}
\begin{table*}[t]
\caption{Performance comparison of intrinsic reward-based methods with only intrinsic rewards on Atari games. The bold font indicates the best scores.}
\label{table-atari}
\begin{center}
\begin{small}
\begin{sc}
\begin{tabular}{lll|llllll}
\hline
Game           & Random   & Human   & ICM     & Disagreement  &RND &NNM & TIR \\ \hline
Alien          & 227.8    & 7127.7  & 374.2  & 316.6  &206.1   &469.3   & \textbf{532.7} \\
Amidar         & 5.8      & 1719.5  & \textbf{356.3}   & 73.3       &   90.8 &198.5 & 27.3  \\
Assault        & 222.4    & 742.0   & 99.4 & 91.6        &308.9 &\textbf{343.6}  & 41.3 \\
Asterix        & 210.0    & 8503.3  & \textbf{2367.6}  & 540.0 &    848.5  &976.3 & 738.0\\
Bank Heist     & 14.2     & 753.1   & 96.8  & 113.7 & 16.6 &16.3 &  \textbf{425.4}    \\
BattleZone     & 2360.0   & 37187.5 & 2794.5 & 3663.3 &  6445.0      &1233.3   &    \textbf{7580.0}    \\
Breakout       & 1.7      & 30.5    & 264.2   & 246.6    &35.5  &106.4 &   \textbf{269.2}   \\
ChopperCommand & 811.0    & 7387.8  & 122.4  & 371.0 &  320.0     &\textbf{1324.0}    &   825.0    \\
Demon Attack   & 107805.0 & 35829.4 &   30.9      &       25.3 & 36.1  &1.3    &    \textbf{44.7}  \\
Freeway        & 0.0      & 29.6    &     0    &    0.2 &\textbf{3.1} &0.1 & 0.3          \\
Gopher         & 257.6    & 2412.5  &    2763.8     &    2456.6    &   1054.8 & 4865.9  &\textbf{5921.6}    \\
Hero           & 1027.0   & 30826.4 &  2559.2     &      2749.7     & 2130.4  &\textbf{3740.5} &2889.0   \\
Jamesbond      & 29.0     & 302.8   & 494.9   & 308.8    & 365.5 &1073.3 & \textbf{4775.0} \\
Kangaroo       & 52.0     & 3035.0  &    557.0     &      514.0   &   412.0 &279.8 &\textbf{1682.8}  \\
Kung Fu Master & 258.5    & 22736.3 &    4226.9     &      2179.5&1135.0   &6006.3  &  \textbf{10844.1}     \\
Ms Pacman      & 307.3    & 6951.6  & 412.7  & 291.0   & 607.2 &476.7 &\textbf{997.7} \\
Pong           & -20.7    & 14.6    &  -9.1       &  -7.9 &   -11.7    &-9.9  &   \textbf{-4.2}     \\
Private Eye    & 24.9     & 69571.3 &    -500    &    \textbf{36.0}    & -997.5 &  0.0    &  0.0   \\
Qbert          & 163.9    & 13455.0 &    \textbf{3558.7}     &      531.3   &  616.0 &  1383.4  &  1349.0  \\
Seaquest       & 68.4     & 42054.7 &   471.9      &      303.6      &  357.8  &272.3 & \textbf{646.2}    \\
Up N Down      & 533.4    & 11693.2 & 15815.1 & 8189.8     & 8736.1 &\textbf{18732.1} &18314.2 \\ \hline
Mean HNS       & 0.0      & 1.0     &   0.767      &   0.637 &    0.284  &0.675  &  \textbf{1.647}     \\
\#SOTA       & N/A      & N/A     &     4    &    1       & 1 &3 & \textbf{12}   \\
\hline
\end{tabular}
\end{sc}
\end{small}
\end{center}
\end{table*}

\begin{figure*}[t] 
    \centering
    \includegraphics[width=1.0\textwidth]{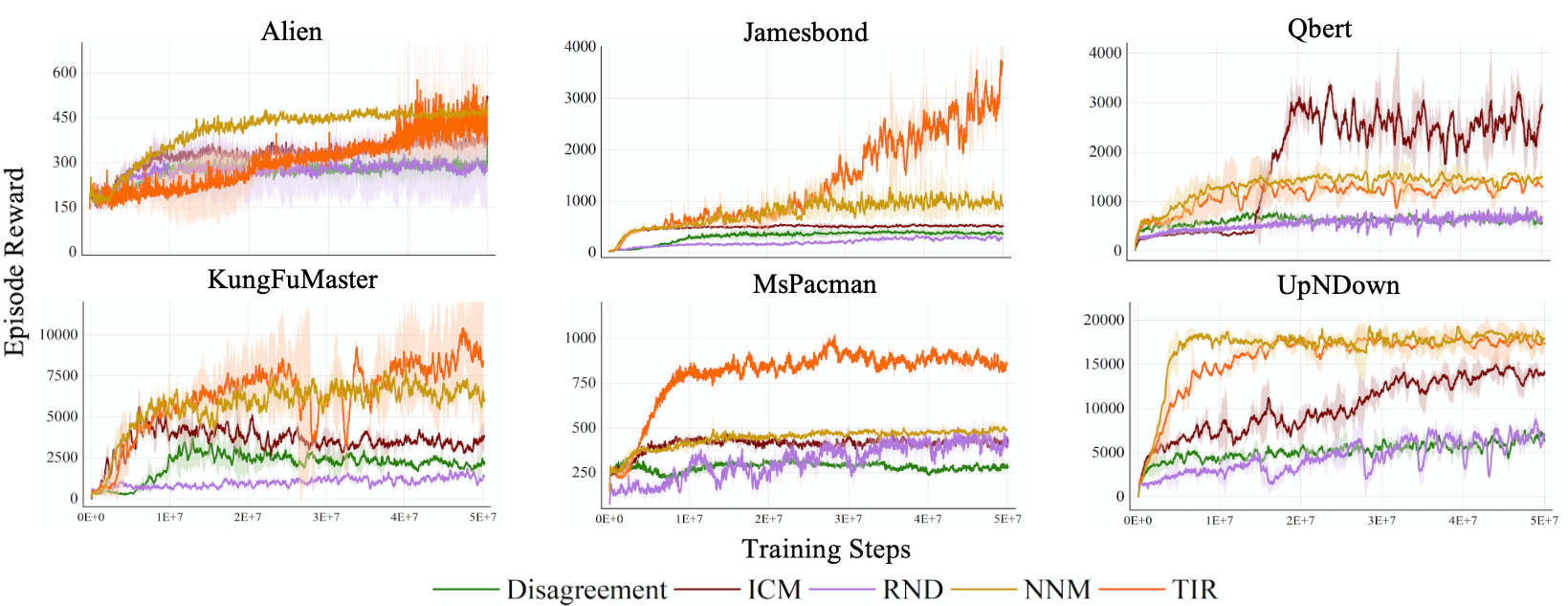}
    \caption{Performance comparison on subsets of Atari games by only using the intrinsic reward. Five competitive methods are compared, including Disagreement, ICM, RND, NNM, and our proposed TIR.} 
    \label{int}
\end{figure*}

\begin{figure*}[t] 
    \centering    \includegraphics[width=1.0\textwidth]{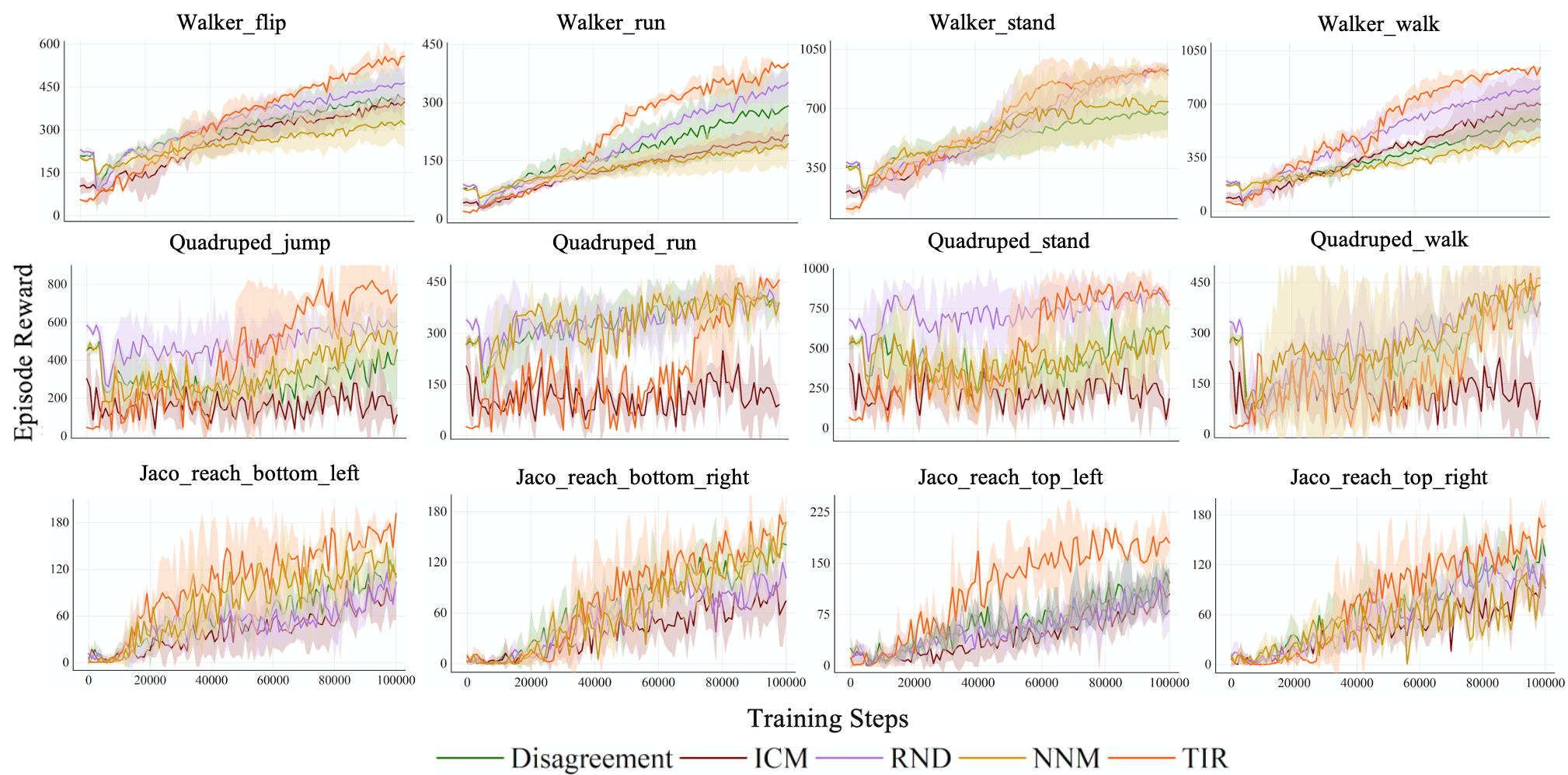}
\caption{\footnotesize{Performance of five competitive intrinsic reward-based methods (Disagreement, ICM, RND, NNM, and TIR) in the fine-tuning phase on DMC environment.}} 
    \label{fig-DMC}
\end{figure*}

\begin{table*}[t]
\caption{Quantitative performance comparison with different pre-training methods on state-based DeepMind Control Suite. The bold font indicates the best results, whereas the underlined font indicates second best results.}
\label{table-DMC}
\renewcommand{\arraystretch}{1.0}
\begin{center}
\begin{small}
\begin{sc}
\resizebox{1.0\linewidth}{!}{
\begin{tabular}{cc|ccccc|cccc|c}
\hline
Domain                      & Task               & ICM            & Disagreement    & RND &NNM            & APT    & ProtoRL & SMM    & DIAYN  & APS             & Ours                           \\ \hline
                            & Flip               & 398±18         & 407±75          & 465±62 &320 ±80        & 477±16 & 480±23  &  \underline{505±26} & 381±17 & 461±24          & \textbf{588±25} \\
                            & Run                & 216±35         & 291±81 & 352±29   &193±45       & 344±28 & 200±15  &  \textbf{430±26} & 242±11 & 257±27          & \underline{400±9}                                 \\
                            & Stand              &  \textbf{928±18}         & 680±107          & 901±8  &742±114         & 914±8  & 870±23  & 877±34 & 860±26 & 835±64          & \underline{923±52}                       \\
\multirow{-4}{*}{Walker}    & Walk               & 696±162         & 595±153          & 814±116   &483±33       & 759±35 & 777±33  & \underline{821±36} & 661±26 & 711±68          & \textbf{938±13}                        \\

\multicolumn{2}{c|}{\textit{Average Performance}}                                   & 560±59            & 494±104    & 633±54 & 435±68   & 624±22 & 582±24 &   \underline{659±31}   & 536±20  & 566±46             & \textbf{713±25}                           \\
\hline
                            & Jump               & 112±4         & 383±265          &  \underline{580±72} &545±107         & 462±48 & 425±63  & 298±39 & 578±46 & 529±42          & \textbf{747±21}                        \\
                            & Run                & 91±29         & 389±61          & 385±47  &390±21        & 339±40 & 316±36  & 220±37 &  \underline{415±28} & 465±37 & \textbf{456±29}                                 \\
                            & Stand              & 184±100         & 628±114          & \textbf{800±54}   &539±76       & 622±57 & 560±71  & 367±42 & 706±48 &  714±50          & \underline{773±19}                                 \\
\multirow{-4}{*}{Quadruped} & Walk               & 99±46         & 384±28          & 392±39 &441±83 & 434±64 & 403±91  & 184±26 & 406±64 &  \textbf{602±86}          & \underline{463±22}                               \\
\multicolumn{2}{c|}{\textit{Average Performance} }                                   & 122±45            & 446±117    & 540±53 & 479±72   & 465±52    & 426±66 &  268±36   & 527±47  &  \underline{578±43}             & \textbf{610±23}                           \\
\hline
                            & Reach bottom left  & 102±47 & 117±17           & 103±17   &110±17       & 88±12  &  \underline{121±22}  & 40±9   & 17±5   & 96±13           & \textbf{191±8}                        \\
                            & Reach bottom right & 75±27          &  142±3           & 101±26 &\underline{165±17}         & 115±12 & 113±16  & 50±9   & 31±4   & 93±9            & \textbf{167±5}                      \\
                            & Reach top left     & 105±29         & 121±17          &  \underline{146±46} &124±18         & 112±11 & 124±20  & 50±7   & 11±3   & 65±10           & \textbf{180±13}                       \\
\multirow{-4}{*}{Jaco}      & Reach top right    & 93±19          & 131±10           & 99±25     &96±8     & \underline{136±5}  & 135±19  & 37±8   & 19±4   & 81±11           & \textbf{187±5}        \\
\multicolumn{2}{c|}{\textit{Average Performance}}                                   & 94±31            &  \underline{128±12}    & 113±29  &124±15 & 113±10    & 124±20 &  44±9   & 20±4  & 84±11             & \textbf{182±8}                           \\
\hline             
\end{tabular}}
\end{sc}
\end{small}
\end{center}
\end{table*}

\begin{figure*}[t] 
    \centering
    \includegraphics[width=1.0\textwidth]{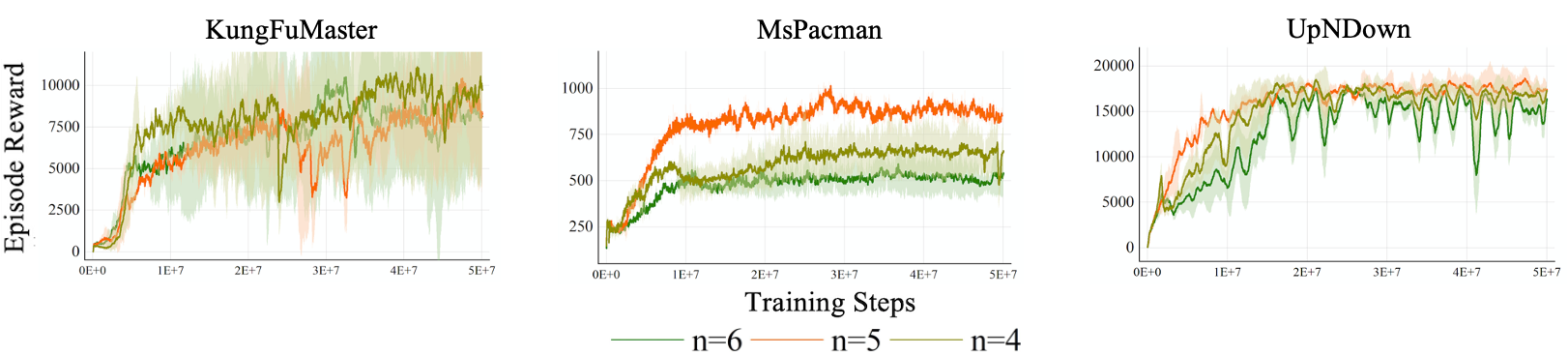}
    \caption{Performance comparison on Atari games between the different number of snapshots by only using the intrinsic reward.} 
    \label{number}
\end{figure*}

\subsection{Experimental Settings}
To conduct a fair quantitative comparison with competing approaches, we evaluate the proposed method using two widely-used benchmarks, including the DeepMind Control Suite (DMC)~\cite{dm_control} and the Atari games~\cite{atari}. We perform the evaluation using two settings: pre-training accompanying fine-tuning, and traditional RL settings. For the Atari environment, we follow the settings of Disagreement~\cite{Disagreemet} and settings of URLB~\cite{urlb} which is a pre-training benchmark for DMC. In all experiments, the number of snapshots is set to 5 and the hyper-parameters $j$ and $k$ are set to 4 and 2, respectively. All comparison experiments use the same settings. We run the experiments across 3 random seeds and employ 50M running steps - equivalent to 200M frames in Atari.

\subsection{TIR surpasses previous curiosity-based methods}
For the Atari environment, we evaluate the performance of agents by only using intrinsic rewards. In other words, they are only motivated by intrinsic rewards for self-supervised exploration. Table~\ref{table-atari} lists the aggregate metrics and scores of four competitive intrinsic rewards (ICM, Disagreement, RND, NNM, and our proposed TIR), on the subset of Atari games. Human and random scores are adopted from~\cite{rainbow}. Following the evaluation settings of previous works \cite{liu2021behavior,yarats2020image,SPR2020data}, we normalize the episode reward as human-normalized scores (HNS), which can be formulated as:
\begin{equation}
    \text{HNS}=\frac{ \text{agent score}-\text {random score}}{ \text{human score}-\text {random score}}
\end{equation}

\#SOTA means the game numbers exceed other methods and the mean HNS is the average of HNS across all games. TIR displays an overwhelming superiority over ICM, Disagreement, RND, and NNM with its highest mean HNS and \#SOTA. Furthermore, the mean HNS score of TIR in the 21 Atari games represents a 114\% relative improvement over other baselines and our intrinsic reward-based method even outperforms the human level in some scenarios, without any feedback from the external environment. 

Figure~\ref{int} provides the learning curves of TIR with four baselines on 6 randomly selected Atari games. Of the 6 games, our method outperforms the Disagreement in 6, the ICM in 5, the RND in 6, and the NNM in 4. TIR has clearly established benefits in terms of performance and learning speed in the majority of games, as seen in the figure. Particularly on the Jamesbond and MsPacman, the convergent episode reward of TIR is more than double that of the other approaches. In brief, our method outperforms other intrinsic reward-based methods in Atari Suite, demonstrating TIR's superiority in providing more accurate intrinsic rewards, which is the critical in self-supervised exploration.

\subsection{TIR surpasses previous pre-training strategies}
The pre-training accompanying fine-tuning paradigm is widely used in the deep learning field and it's also the key technique to improve sample efficiency for DRL~\cite{urlb}. DMC environment encompasses a variety of difficult domains and tasks of different complexity, and it is built on robot simulations, which makes modeling the robot dynamics inherently tough. We conduct extensive experiments on the DMC to demonstrate the superiority of the TIR based on the pre-training paradigm. We evaluate all baselines and TIR from the easiest to the most difficult domains (Walker, Quadruped, and Jaco), which each have four tasks. During the pre-training phase, an agent is trained for 2 million steps with only intrinsic reward and only 100k steps with extrinsic reward during the fine-tuning stage.

Figure~\ref{fig-DMC} plots the learning curves (fine-tuning phase) of the aforementioned four methods in all tasks. 
As shown in the figure, TIR not only provides a faster convergence speed than other intrinsic reward-based methods, but the convergence result also significantly outperforms baselines. 
We suppose the improvements in the convergence speed rest from historical information provided by multiple snapshots. Additionally, we compare the final scores and standard deviations of TIR, with the intrinsic reward-based methods (ICM, Disagreement, RND, NNM, APT~\cite{liu2021behavior}) and other pre-training strategies (ProtoRL~\cite{prot}, SMM~\cite{lee2019efficient}, DIAYN~\cite{eysenbach2018diversity}, APS~\cite{liu2021aps}). The quantitative results in Table~\ref{table-DMC} clearly show that TIR achieves state-of-the-art results in 8 of 12 tasks and the average performances in all domains surpass all the other methods. Especially, in the hardest domain, our approach improves average performance by 42.0\%, which demonstrates TIR's great potential to improve model performance and robustness in the pre-training paradigm. 

To conclude, large-scale experimental results successfully certify our analysis that the prediction error may be noisy and inaccurate. Disagreement~\cite{Disagreemet} and NNM~\cite{chen2022nuclear} seek to address the issue by introducing multiple models to access model uncertainty and generate the lower-variance estimate. However, due to different initial parameters and stochasticity, it is extremely difficult for models to achieve an agreement, resulting in a worse performance than ICM. Instead, we reuse previous snapshots as the historical knowledge to avoid additional stochasticity (like initial parameters), which is also employed by Averaged-DQN~\cite{anschel2017averaged} to estimate the Q value. As a result, our curiosity formulation outperforms the others, and TIR shows significant performance improvement without any additional training costs. There is also convincing evidence showing that when pre-trained with the prediction model, the pre-trained policy for 500k steps outperforms the pre-trained policy for 2M steps in downstream tasks, indicating that temporal information should be considered~\cite{yuan2022euclid}.

\subsection{Further Analysis}
\label{furtheranalysis}
In this subsection, we main investigate the robustness of TIR with different settings.
\subsubsection{Tolerance of Ensemble Size}
The number of snapshots, which is the size of ensemble, is a critical TIR setting. To investigate the sensitivity to ensemble size, we compare TIR performance with a different number of snapshots of ensemble size $n$. The performance comparison is shown in Figure~\ref{number} and we can observe that TIR with different settings all show promising performance, demonstrating the robustness. To balance the performance and computation costs, in this article, we set the number of snapshots as 5.

\subsubsection{Tolerance of Noise}
\begin{figure}[ht]
\begin{center}
\centerline{\includegraphics[width=\columnwidth]{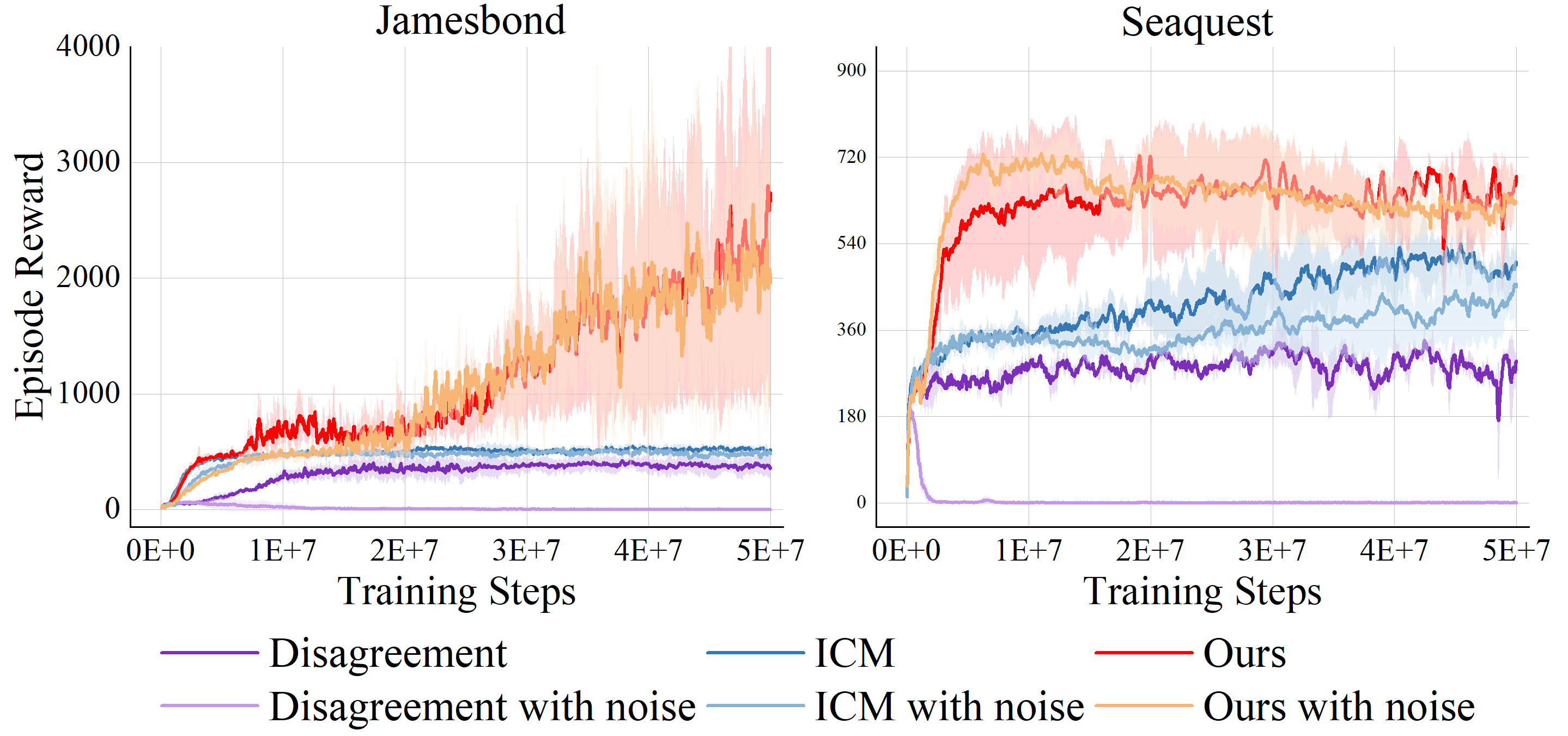}}
\caption{The performance comparison between baselines and TIR with the contamination of noise in Atari Suite.}
\label{noise}
\end{center}
\end{figure}
As previously stated, predicting the future is extremely difficult, particularly in the presence of noise. We introduce 0.25 standard deviation Gaussian noise to the feature vector in order to verify the robustness of baselines and TIR~\cite{dean2020see}. Figure~\ref{noise} plots the performance of agents perturbed with and without noise across two games. We can find that the performance of Disagreement degrades apparently in both two games and the performance of ICM also degrades. On the contrary, TIR with the noise still shows great performance compared to baselines, which proves the robustness of TIR over Disagreement and ICM.

\subsubsection{Sensitivity Analysis for hyper-parameters}

\begin{table}[ht]
\footnotesize
\caption{Performance comparison among different hyper-paramter weights of $j$ across three games of Atari.}
\centering
\label{j}
\begin{tabular}{c|c|c|c}
\hline \diagbox{Method}{Game} & Jamesbond & Gopher & UpNDown \\
\hline Random & 29.0 & 257.6 & 533.4 \\
\hline Human & 302.8 & 2412.5 & 11693.2 \\
\hline Disagreement & 308.8 & 2456.6 & 8189.8 \\
\hline ICM & 494.9 & 2763.8 & 15815.1 \\
\hline RND & 365.5 & 1054.8 & 8736.1 \\
\hline NNM & 873.3 & 4865.9 & 18732.1 \\
\hline TIR$_{j=4}$ & \underline{4775.0} & \underline{5921.6} & 18314.2 \\
\hline TIR$_{j=5}$ & \textbf{7595.0} & \textbf{5966.4} & \underline{19142.3} \\
\hline TIR$_{j=6}$ & 1779.0& 5194.6 & \textbf{21396.1} \\
\hline TIR$_{j=7}$ & 1804 & 5295.6 & 12374.1 \\
\hline
\end{tabular}
\end{table}

In TIR, there is a trade-off between updating the snapshots too often and updating the snapshots too slowly and hyper-parameter $j$ denotes the temporal gap between different snapshots.
To analyze the sensitivity, we evaluate TIR with different settings hyper-parameter $j$. As illustrated in Table~\ref{j}, all values of the hyper-parameter $j$ between 4 and 7 provide a significant advantage over both baselines and human scores, evidently proving the stability and robustness of TIR. Additionally, we can observer the wild variation in performance between different value of $j$ in Jamesbond scenario. It's because that when $j$ is too big, the snapshots parameters may be very different and the snapshot ensemble can be regarded as a variant of the ensemble in Disagreement which limits the performance of TIR. 

\begin{figure}[ht]
\begin{center}
\centerline{\includegraphics[width=\columnwidth]{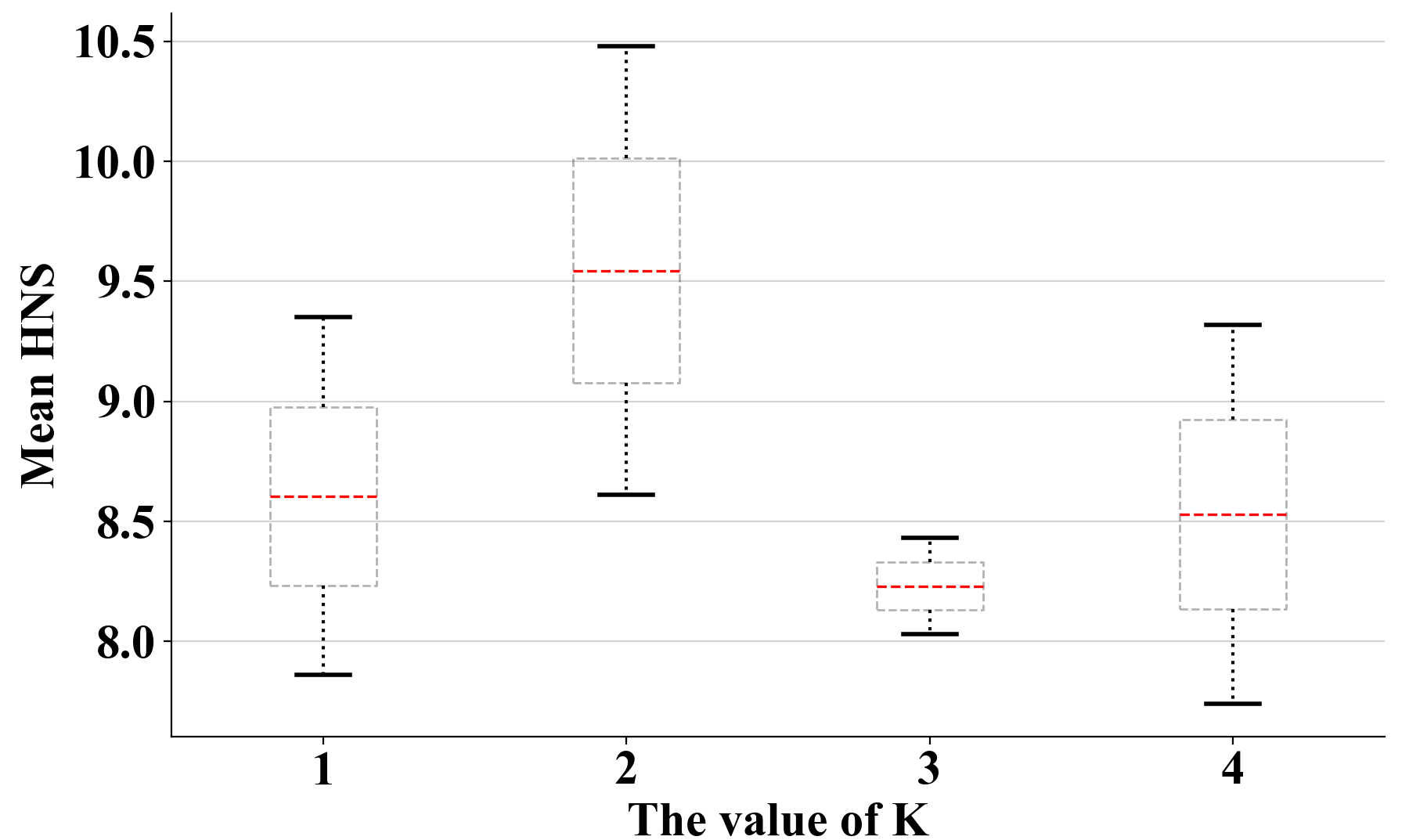}}
\caption{The effect of the weight factor $k$ on HNS value.}
\label{k}
\end{center}
\end{figure}

The hyper-parameter $k$ can adjust the weight of singular values and the bigger $k$ is, the smaller the weight is. We set $k = \{1,2,3,4\}$ and compare the HNS values under different $k$ in three Atari games (Breakout, Gopher and Jamesbond). In particular, when $k=1$, TIR is equivalent to evaluating intrinsic rewards with the raw nuclear norm. From Figure~\ref{k}, we can see that when $k=2$ TIR has produced superior results with a higher HNS value. Furthermore, under all different $k$ values, TIR greatly outperforms humans in both three games. Such evidence demonstrates the TIR shows the robustness to different settings of hyper-parameter $k$.

\subsubsection{Selection of Nuclear Norm}
\begin{table}[ht]
\footnotesize
\caption{Performance comparison between baselines and different settings of TIR. The agents are trained with only intrinsic rewards.}
\centering
\label{norm}
\begin{tabular}{c|c|c|c}
\hline \diagbox{Method}{Game} & Jamesbond & MsPacman & UpNDown \\\hline
ICM & 494.9 & 412.7 & 15815.1 \\\hline
Disagreement & 308.8 & 291.0 & 8189.8 \\\hline
RND & 365.5 & 607.2 & 8736.1 \\\hline
NNM & 1073.3 & 476.7 & \textbf{18732.1} \\\hline
Variance & 530.5 & 328.5 & 2910.8 \\\hline
$\ell^1$ norm & 963.0 & 85.3 & 16642.6 \\\hline
$\ell^2$ norm & 2270,0 & 329.0 & 11178.5 \\\hline
$\ell^F$ norm & 1255.0 & 337.3 & 12328.6 \\\hline
Nuclear norm & 2333 & 364.9 & 10894.0 \\\hline
Fix\_weighted\_NN & \underline{3421.5} & \underline{697.3} & 17123.0 \\\hline
TIR & \textbf{4775.0} & \textbf{997.7} & \underline{18314.2} \\
\hline
\end{tabular}
\end{table}
We propose that the superior performance of TIR can be attributed to two major factors: its ability to make better use of historical knowledge with multiple snapshots and the nuclear norm's more accurate assessment of curiosity compared to other norms. To verify this, we conducted controlled experiments to evaluate TIR's performance using various norms, including $\ell^1$, $\ell^2$, $\ell^F$ norm, etc. We also compared TIR's performance with the nuclear norm and the fixed weighting nuclear norm, where $k$ is fixed and is set to 2. As shown in Table~\ref{norm}, in all three games, the agents' scores in all different settings of TIR demonstrate promising performance. By comparing the nuclear norm and fixed weighted nuclear norm settings, we can validate our hypothesis that inaccurate predictions result in large singular values of the prediction matrix and inaccurate rewards, and we need to avoid the influence of under-fitting snapshots with the adaptive weighted nuclear norm. For this reason, in Jamesbond and MsPacman, the scores of TIR exhibit significant advantages over both baselines and other settings, indicating the importance of temporal information and the effectiveness of the variational weighting mechanism.

\section{Conclusion}
\label{conclusion}
To address the challenge of sparse rewards, we propose a simple yet effective intrinsic reward for DRL. Specifically, we train a self-supervised prediction model and maintain a set of snapshots of the prediction model parameters for free. We further utilize the nuclear norm to evaluate the temporal inconsistency between the predictions of different snapshots, which can be further deployed as the intrinsic reward. Moreover, the variational weighting mechanism is proposed to adaptively assign weights to different snapshots. Our solution does not incur additional training costs, while maintaining higher noise tolerance. In two widely-used environments, TIR shows a promising advantage compared to other competing intrinsic rewards-based methods. In the future, we will further investigate improving the prediction performance with more trajectory information instead of using only the information from one transition.


\bibliography{ref.bib}
\bibliographystyle{IEEEtran}

\end{document}